\documentclass[review]{elsarticle}
\usepackage{lineno,hyperref}
\modulolinenumbers[1]
\journal{Fifth EAGE Digitalization Conference \& Exhibition, 24-26 March 2025, United Kingdom}
\date{}
\usepackage[utf8]{inputenc}
\usepackage{amsmath,amsthm,amsfonts}
\usepackage{enumitem}
\usepackage{graphicx}
\usepackage{color}
\usepackage{float}
\usepackage{placeins}
\usepackage{multirow}
\usepackage{subcaption}
\usepackage{algorithm}
\usepackage{algpseudocode}
\usepackage{amsmath}
\usepackage{afterpage}
\usepackage{geometry}
\usepackage{float}

\graphicspath{ {./}}
\begin{document}
\makeatletter
\def\ps@pprintTitle{%
  \let\@oddhead\@empty
  \let\@evenhead\@empty
  \def\@oddfoot{\hfill\footnotesize\itshape
       \ifx\@journal\@empty\else\@journal\fi\hfill}%
  \let\@evenfoot\@oddfoot}
\makeatother

\begin{frontmatter}

\title{Well Log-Guided Synthesis of Subsurface Images from Sparse Petrography Data Using cGANs}

\author[address1]{Ali Sadeghkhani}
\author[address2]{A. Assadi}
\author[address1]{B. Bennett}
\author[address1]{A. Rabbani\corref{mycorrespondingauthor}}
\cortext[mycorrespondingauthor]{Corresponding author}
\ead{a.rabbani@leeds.ac.uk | rabarash@gmail.com}

\address[address1]{School of Computer Science, University of Leeds, Leeds, UK}
\address[address2]{TEC}

\begin{abstract}
Pore-scale imaging of subsurface formations is costly and limited to discrete depths, creating significant gaps in reservoir characterization. 
To address this, we present a conditional Generative Adversarial Network (cGAN) framework for synthesizing realistic thin section images of 
carbonate rock formations, conditioned on porosity values derived from well logs. The model is trained on 5,000 sub-images extracted from 15 petrography samples over a depth interval of 1992-2000m, the model generates  geologically consistent images across a wide porosity range (0.004-0.745), achieving 81\% accuracy within a 10\% margin of target porosity values. 
The successful integration of well log data with the trained generator enables continuous pore-scale visualization along the wellbore, bridging gaps between discrete core sampling points and providing valuable insights for reservoir characterization and energy transition applications such as carbon capture and underground hydrogen storage.
\end{abstract}

\begin{keyword}
Conditional Generative Adversarial Networks \sep Digital rock \sep Porous media \sep Porosity control \sep Deep learning \sep Subsurface imaging \sep Thin sections
\end{keyword}
\end{frontmatter}

\section{Introduction}
The accurate prediction of physical properties in porous materials is crucial for geoscience exploration and various engineering applications. The three-dimensional microstructure of these materials defines their macroscopic properties, including porosity, permeability, and fluid flow behavior. However, subsurface fluid flow studies face persistent data scarcity challenges, particularly in visual pore-scale data, which is a foundation for digital rock analysis. While techniques like scanning electron microscopy (SEM) and X-ray computed tomography offer high-resolution imaging capabilities, they are limited by cost, accessibility, and field of view constraints \citep{Hemes2015, Li2018}. These images are typically available only at specific depths and wells, creating significant gaps in understanding of full-depth formations. To address this challenge, we present a novel approach using Conditional Generative Adversarial Networks (cGANs) to generate representative thin section images across the full depth of subsurface formations, using limited existing images and corresponding porosity profiles.

\section{Method and Theory}

Generative Adversarial Neural Networks (GANs) constitute a novel deep learning framework for generative modeling, introduced by \cite{Goodfellow2014}. The framework consists of two competing neural networks, a generator and a discriminator. The generator network ($G$) tries to learn the underlying data distribution by creating new, synthetic samples. Simultaneously, the discriminator network ($D$) acts as a classifier, aiming to distinguish between real data and samples generated by $G$. This ongoing competition refines the capabilities of both networks. The generator progressively improves its ability to generate realistic samples, while the discriminator develops its skill in differentiating real from artificial data.

Previous applications of GANs in porous media reconstruction have shown promising results in capturing complex geological structures. Early implementations of Deep Convolutional GANs successfully reconstructed three-dimensional porous samples across various rock types while preserving key physical properties \citep{Mosser2017, Mosser2018}. Subsequent developments enhanced these capabilities through advanced architectures that could process high-resolution volumetric data and generate 3D structures using 2D training data \citep{Kench2021, Chi2024}, contributing to the growing body of work in digital rock physics applications and establishing GANs as an effective tool for subsurface characterization.

Our methodology employs a conditional GAN (cGAN) architecture, developed by \cite{Mirza2014}, to generate pore-scale subsurface images from sparse geological thin section data of carbonate rock formations. The network architecture incorporates porosity as a conditional parameter, where the generator synthesizes images from random noise conditioned on porosity values, while the discriminator evaluates both the image and its corresponding porosity condition. This conditioning mechanism enables the network to learn the relationship between porosity and structural features, allowing for targeted image generation. The approach effectively addresses subsurface data scarcity by generating representative images based on porosity data from well logs, thereby providing additional information for rock typing and interpretation in areas with limited direct imaging data.

The cGAN architecture consists of a generator with 6 transposed convolutional layers using LeakyReLU activation and a discriminator with 5 convolutional layers. The training dataset comprises 5,000 sub-images of size $256\times256$ pixels that were extracted from 15 petrography images of single epoxy-stained thin-section carbonate samples (interval 1992-2000m). Each image underwent analysis for porosity using threshold-based segmentation in HSV color space. The images were systematically categorized based on porosity ranges into 10 classes, and data augmentation was employed to achieve a balanced distribution across all classes for training.

The training process employed binary cross-entropy loss with the Adam optimizer (learning rate: $2\times10^{-4}$) over 100 epochs. Both networks were simultaneously trained, the generator to produce increasingly realistic images that match the conditional parameters, and the discriminator to improve its ability to distinguish between real and generated samples while considering the specified condition. This adversarial training process, combined with the conditional input, enables the model to generate geologically realistic images with controlled physical properties.

\section{Results}

\begin{figure}[!htb]
 \centering
 \includegraphics[width=0.8\textwidth]{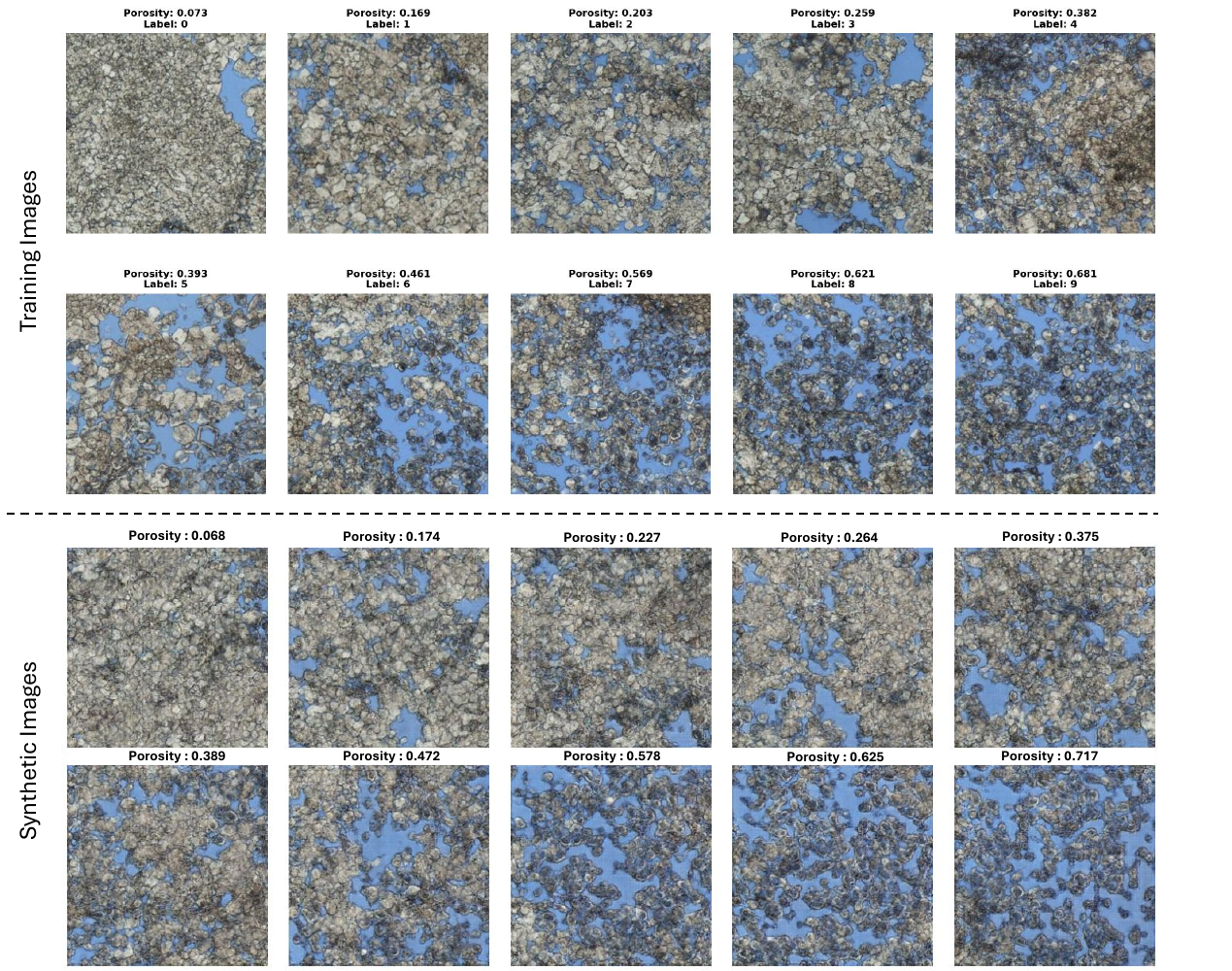}
 \caption{Comparison between training images (top two rows) and synthetic images (bottom two rows) generated by the cGAN model. Each image, with dimensions of 256 × 256 pixels, is labeled with its corresponding porosity value, demonstrating the model's ability to generate geologically consistent images across various porosity ranges.}
 \label{fig:comparison}
\end{figure}

The results demonstrate the model's ability to capture and reproduce complex geological features such as intra- and inter-particle porosities across a wide range of porosity values (0.004-0.745). Visual comparison between the training and synthetic images (Figure \ref{fig:comparison}) reveals successful reproduction of key carbonate rock features, including pore network architectures and grain boundary relationships. The synthetic images maintain geological realism while matching the target porosity conditions, as evidenced by the consistent pore size distributions and spatial arrangements across both training and generated samples.

\begin{figure}[ht]
 \centering
 \includegraphics[width=0.7\textwidth]{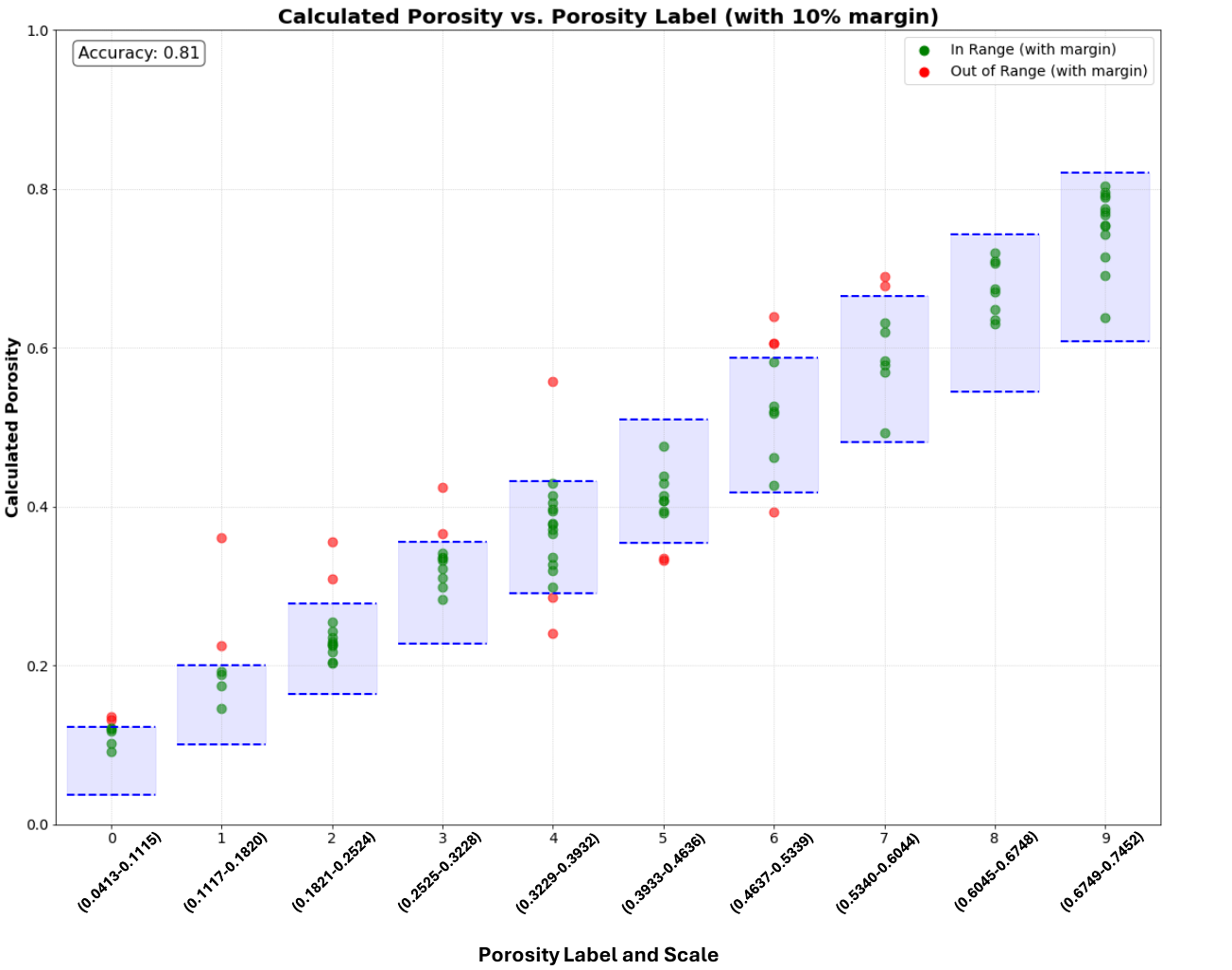}
 \caption{Quantitative validation of generated images showing calculated porosity versus porosity labels (with corresponding porosity ranges shown on x-axis). Green points indicate samples within the acceptable 10\% margin of their target porosity range, while red points show outliers. The model achieves 81\% accuracy within the specified margin.}
 \label{fig:accuracy}
\end{figure}

Quantitative analysis of the generated samples reveals high accuracy, with 80\% of the generated porosity values falling within a 10\% margin of the target ranges, as shown in Figure \ref{fig:accuracy}. This technique addresses data scarcity in subsurface modeling by generating representative images based on porosity data from well logs and provides additional information for interpretations and rock typing in areas with limited direct imaging data.

\begin{figure}[H]
 \centering
 \includegraphics[width=0.6\textwidth]{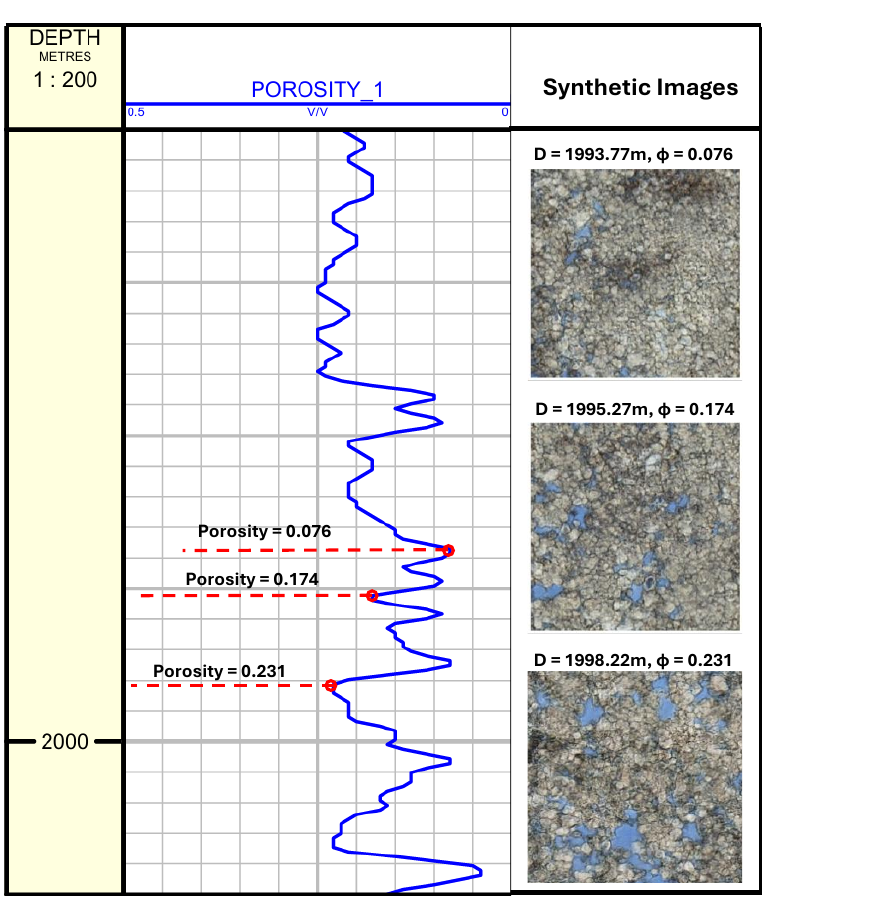}
 \caption{Well log-guided image synthesis showing the porosity log (blue curve) and corresponding synthetic thin section images generated at selected depths. The generated images demonstrate the model's ability to produce geologically consistent representations based on well-log porosity values, enabling continuous visualization of pore-scale features along the wellbore.}
 \label{fig:well_log}
\end{figure}

To demonstrate the practical application of our trained model, we generated synthetic thin section images at various depths using porosity values obtained from well log data (Figure \ref{fig:well_log}). The generated images show consistent geological features corresponding to the porosity variations observed in the well log. This capability enables continuous visualization of pore-scale features along the wellbore, effectively bridging the gaps between actual core sampling points and providing insights into the spatial variation of rock properties.

\section{Conclusions}
This approach helps reduce subsurface characterization uncertainties in implementing energy transition technologies like carbon capture and storage and underground hydrogen storage. The cGAN model demonstrates robust performance in generating geologically consistent thin section images across a wide range of porosity values (0.004-0.745), achieving 80\% accuracy within a 10\% margin of target porosity values. The successful integration of well log data with the trained generator enables continuous visualization of pore-scale features along the wellbore, effectively bridging the gaps between discrete core sampling points. This capability significantly enhances our understanding of spatial variations in rock properties and provides valuable insights for reservoir characterization.

The study establishes a foundation for investigating additional parameters and on-demand 3D generation of pore-scale images across the full depth of studied subsurface formations. Future work could extend this methodology to incorporate multiple conditioning parameters such as permeability and mineralogy, further enhancing the utility of this approach for comprehensive reservoir characterization. The developed framework shows promising potential for reducing the cost and time associated with subsurface characterization while maintaining geological consistency and quantitative accuracy in the generated results.

\end{document}